\documentclass{article}
\usepackage{graphicx}
\usepackage{subcaption}

\usepackage{tabularx}
\usepackage{array}
\usepackage{xcolor}
\usepackage{geometry}

\usepackage{multirow}
\usepackage{amsmath}
\usepackage[T1]{fontenc}
\usepackage[utf8]{inputenc}
\usepackage{multicol}
\usepackage{xurl}
\usepackage[hidelinks]{hyperref}

\title{Tiny, On-Device Decision Makers with the MiniConv Library}
\author{Carlos Purves \\
University of Cambridge \\
\texttt{cp614@cam.ac.uk}}
\date{2024}

\begin{document}

\maketitle

\begin{abstract}
	    Reinforcement learning (RL) has achieved strong results, but deploying visual policies on resource-constrained edge devices remains challenging due to computational cost and communication latency. Many deployments therefore offload policy inference to a remote server, incurring network round trips and requiring transmission of high-dimensional observations. We introduce a split-policy architecture in which a small on-device encoder, implemented as OpenGL fragment-shader passes for broad embedded GPU support, transforms each observation into a compact feature tensor that is transmitted to a remote policy head. In RL, this communication overhead manifests as closed-loop decision latency rather than only per-request inference latency. The proposed approach reduces transmitted data, lowers decision latency in bandwidth-limited settings, and reduces server-side compute per request, whilst achieving broadly comparable learning performance by final return (mean over the final 100 episodes) in single-run benchmarks, with modest trade-offs in mean return. We evaluate across an NVIDIA Jetson Nano, a Raspberry Pi 4B, and a Raspberry Pi Zero 2 W, reporting learning results, on-device execution behaviour under sustained load, and end-to-end decision latency and scalability measurements under bandwidth shaping. Code for training, deployment, and measurement is released as open source.
\end{abstract}


\section{Introduction}

Reinforcement learning (RL) has achieved strong results in domains such as Go~\cite{Silver16} and Atari~\cite{Mnih13}, but deploying visual policies on edge devices remains challenging due to their compute and communication requirements~\cite{alajlan2022tinyml}. The gap between simulation performance and real-world deployment remains substantial~\cite{SimRealComb}, particularly where decisions must be made at low latency and under tight power and memory budgets.

A common deployment pattern is to execute the policy on a remote server. This introduces \emph{decision latency} (wall-clock time from observation availability to action receipt) through network round trips and can add jitter under congestion. Transmitting raw visual observations can also be bandwidth-intensive and may raise privacy concerns~\cite{liu2021machine}.

This paper proposes a split-policy architecture that moves early visual feature extraction to the client device. A lightweight on-device encoder produces a compact feature tensor that is transmitted to a remote policy head, trading a small on-device compute cost for reduced communication, lower decision latency in bandwidth-limited settings, and reduced server-side compute per request.

We realise this architecture using MiniConv: a library of small convolutional encoders designed to compile cleanly to OpenGL fragment shaders. The key differentiator is that the encoder is expressed as a small sequence of fragment-shader passes designed for the constraints of embedded OpenGL implementations. This enables on-device execution on widely supported embedded GPUs and integrates the encoder with existing graphics pipelines. We evaluate MiniConv across three representative devices (NVIDIA Jetson Nano, Raspberry Pi 4B, Raspberry Pi Zero 2 W), reporting learning performance under visual observations, sustained on-device execution behaviour, end-to-end decision latency under bandwidth shaping, and server scalability.

Our contributions are: (i) MiniConv and an OpenGL deployment pathway for small visual encoders, (ii) a split-policy RL pipeline that transmits compact visual features rather than raw frames, and (iii) an empirical evaluation and open-source measurement tooling spanning learning, latency, and device resource behaviour.

\section{Related Work}

Real-world applications of reinforcement learning (RL) have rapidly expanded in recent years~\cite{kiran2021deep}. One of the most prominent areas~\cite{piazza2019century} of application is in robotics~\cite{yu2022dexterous}, where RL has enabled robots to learn complex motor skills~\cite{andrychowicz2020learning}, adaptive navigation~\cite{tai2017virtual}, and autonomous manipulation of objects in dynamic environments~\cite{matas2018sim}.

There is extensive work on running neural networks on edge devices. Approaches range from specialist additional tensor processing hardware, such as the Coral USB Accelerator\footnote{\url{https://coral.ai/products/}}, to model pruning~\cite{zhu2017prune} and specialist model designs~\cite{RAY20221595}.

MobileNet~\cite{Howard17} is a family of architectures designed for efficient inference on mobile and low-power devices. It achieves a favourable accuracy--efficiency trade-off using depthwise separable convolutions and related design choices that reduce parameter count and multiply--accumulate operations.

Subsequent variants introduced further efficiency improvements. MobileNetV2 employed inverted residual blocks with linear bottlenecks, whilst MobileNetV3 combined these with neural architecture search and squeeze-and-excitation to produce more efficient backbones. These optimised designs can nevertheless underperform larger architectures on visually complex tasks.

Complementary approaches reduce an existing model’s compute and memory footprint. Model compression methods include pruning~\cite{zhu2017prune,blalock2020state}, quantisation~\cite{paupamah2020quantisation}, and knowledge distillation~\cite{gou2021knowledge}, and are surveyed in~\cite{2018arXiv180203494H}.

More directly related to split-policy execution, several systems partition deep neural network inference between end devices and the edge or cloud to optimise latency and resource usage under bandwidth constraints. Neurosurgeon~\cite{DBLP:conf/asplos/KangHGRMMT17} selects partition points in DNNs to balance device computation against transmission cost, whilst Edge Intelligence~\cite{DBLP:conf/sigcomm/LiZC18} explores on-demand co-inference with device--edge synergy. Teerapittayanon \emph{et al.}~\cite{DBLP:conf/icdcs/Teerapittayanon17} consider distributed DNN execution across end devices, edge servers, and the cloud. MiniConv is complementary: it applies a similar division of labour to RL policies, emphasising wide hardware support through OpenGL shader execution and transmitting compact feature representations rather than raw observations. This work evaluates the resulting trade-offs in decision latency, scalability, and device resource pressure.

\section{Implementation}

The \emph{MiniConv library} provides small, composable encoder blocks designed to compile cleanly to OpenGL fragment shaders, respecting practical constraints such as texture binding and sampling limits. In this paper we instantiate MiniConv encoders with $K$ output channels (specifically $K=4$ and $K=16$) and train them end-to-end together with a downstream policy in PyTorch. At deployment, only the MiniConv encoder runs on-device (via OpenGL), producing a $K$-channel feature tensor per frame; only this tensor is transmitted to the server-side policy head. MiniConv is a \emph{library} rather than a single fixed architecture: $K$ and block compositions can be varied to meet device and bandwidth constraints.

\begin{figure}[t]
    \centering
    \begin{subfigure}{0.45\textwidth}
        \centering
        \includegraphics[width=\linewidth]{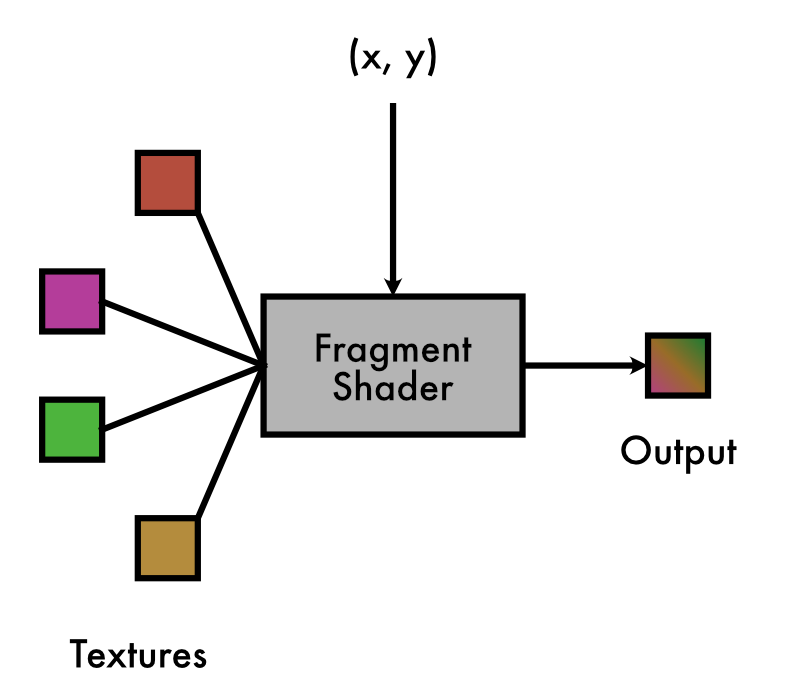}
        \caption{Fragment shader input/output.}
        \label{fig:glfrag:1:short}
    \end{subfigure}
    \hfill
    \begin{subfigure}{0.45\textwidth}
        \centering
        \includegraphics[width=\linewidth]{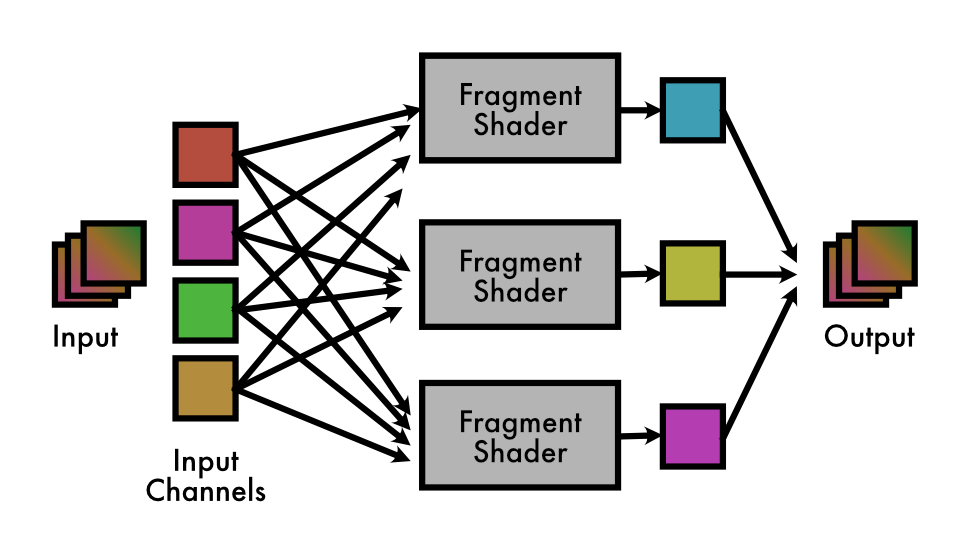}
        \caption{Mapping CNN layers to shader passes.}
        \label{fig:glfrag:2:short}
    \end{subfigure}
    \caption{OpenGL fragment shaders can implement convolution and pooling by sampling input textures and writing output textures.}
    \label{fig:glfrag:short}
\end{figure}

We deploy the on-device encoder using OpenGL fragment shaders, which compute each output pixel as a function of one or more input textures and are widely supported across embedded GPUs. This execution model maps naturally to convolution and pooling: a shader samples a neighbourhood of an input texture and writes an output texture, as illustrated in Figure~\ref{fig:glfrag:short}. MiniConv exploits this mapping whilst respecting the practical limits of low-cost devices. For example, on the Raspberry Pi Zero 2 W, fragment shaders can sample from a maximum of eight bound textures, and each shader is subject to a finite sampling budget (64 texture samples in our deployment). Since each shader pass outputs four channels (RGBA), encoders with larger $K$ are implemented via multiple passes. These constraints inform the choice of kernel sizes, channel packing, and layer compositions used by MiniConv.

\section{Evaluation}

Deploying split-policy RL on edge devices requires that the on-device encoder preserves policy performance whilst respecting strict compute, memory, and power constraints. We therefore organise the evaluation around eight practical questions:
\begin{description}
    \item[\textbf{Q1}] Does a split-policy architecture match the learning performance of a conventional Full-CNN baseline under visual observations?
    \item[\textbf{Q2}] Does the compressed on-device representation retain sufficient task-relevant information to support high-return behaviour?
    \item[\textbf{Q3}] How do per-frame inference latency and variability change under sustained on-device execution?
    \item[\textbf{Q4}] What memory footprint does on-device inference impose, and how much RAM headroom remains for other tasks?
    \item[\textbf{Q5}] What is the effect of sustained inference on device thermal state and throttling behaviour?
    \item[\textbf{Q6}] At what link bandwidth does split inference reduce end-to-end decision latency relative to transmitting full observations?
    \item[\textbf{Q7}] On low-power devices, how does OpenGL shader execution compare to a CPU implementation in throughput and stability?
    \item[\textbf{Q8}] How do power limits and power consumption affect inference throughput and stability?
\end{description}
We address these questions through learning experiments on visual control tasks, on-device execution benchmarks, and end-to-end measurements of decision latency and server scalability under bandwidth constraints.

\subsection{Learning}

We evaluate MiniConv encoders on two MuJoCo locomotion tasks (\emph{Walker2d}, \emph{Hopper}) and the classic control \emph{Pendulum} task under visual observations. We train \emph{Walker2d} with PPO~\cite{schulman2017proximalpolicyoptimizationalgorithms}, \emph{Hopper} with SAC~\cite{haarnoja2018softactorcriticoffpolicymaximum}, and \emph{Pendulum} with DDPG~\cite{lillicrap2015continuouscontroldeepreinforcement}, selected based on preliminary stability under pixel observations and standard practice in Stable-Baselines3 for the respective tasks. Unless otherwise stated, \emph{Walker2d} and \emph{Hopper} are trained for 2{,}000 episodes and \emph{Pendulum} for 1{,}000 episodes. Because algorithms differ across tasks, cross-task comparisons are not meaningful; we therefore focus on within-task comparisons between encoders. Results are reported for a single run per condition (fixed seed), and variance across seeds is not yet characterised.

\subsubsection*{Algorithms and baselines}
\noindent Table~\ref{tab:algs:short} summarises the learning algorithm used for each task.
\begin{table}[t]\centering
\caption{Algorithms used for each visual control task.}
\label{tab:algs:short}
\begin{tabularx}{\linewidth}{l l X}\hline
\textbf{Task} & \textbf{Algorithm} & \textbf{Selection rationale} \\\hline
Walker2d-v4 & PPO & On-policy baseline that trained without collapse under pixel observations in our experimental configuration. \\
Hopper-v4   & SAC & Common off-policy baseline for continuous control that trained without collapse under pixel observations in our experimental configuration. \\
Pendulum-v1 & DDPG & Lightweight deterministic baseline that trained without collapse for Pendulum under pixel observations in our experimental configuration. \\
\hline\end{tabularx}\end{table}

For each task, the Full-CNN baseline corresponds to the default convolutional feature extractor used by Stable-Baselines3~\cite{stable-baselines3} for image observations (\texttt{CnnPolicy}). The MiniConv conditions replace only this observation encoder (with $K \in \{4,16\}$ output channels); the downstream policy and value networks are unchanged. Within each task, all other training settings are held fixed across encoder variants, including optimiser configuration and the algorithm-specific parameters that most directly affect learning dynamics (PPO rollout and clipping; SAC replay buffer and entropy regularisation; DDPG exploration noise and target updates). Unless otherwise stated, these settings follow the Stable-Baselines3 defaults. The split-policy architecture does not assume a particular RL algorithm; however, we do not present a controlled cross-algorithm comparison, and results should be interpreted as within-task evidence that encoder partitioning can be compatible with learning under multiple common RL algorithms.

All experiments use \emph{visual} observations: the environment state is provided as rendered RGB frames rather than low-dimensional state vectors.

We use Gymnasium~\cite{towers2024gymnasiumstandardinterfacereinforcement} with pixel observations (\texttt{render\_mode}=\allowbreak \texttt{rgb\_array}). We consider \emph{Walker2d-v4} and \emph{Hopper-v4} (MuJoCo~\cite{todorov2012mujoco}), and \emph{Pendulum-v1} (Classic Control). For MuJoCo tasks, the OpenGL backend is fixed per run (\texttt{MUJOCO\_GL=\allowbreak egl} or \texttt{MUJOCO\_GL=\allowbreak osmesa}) to ensure deterministic rendering. Cameras are fixed: the MuJoCo tracking camera (\texttt{track}) for \emph{Walker2d-v4}/\emph{Hopper-v4} and the default static camera for \emph{Pendulum-v1}.

Each step renders a 100$\times$100 RGB uint8 frame, which is cropped to 84$\times$84 (random crop during training; deterministic centre crop during evaluation) without resizing interpolation. RGB frames are converted to float32 and normalised to [0,1] using Stable-Baselines3’s default image normalisation (\texttt{normalize\_images=True}). Temporal information is provided by stacking three consecutive frames, yielding a 9$\times$84$\times$84 channel-first tensor after transposition with \texttt{VecTransposeImage}. For deployment and bandwidth analyses only, an opaque alpha channel (255) is appended at the OpenGL upload boundary to form RGBA; training uses RGB only.

The wrapper stack applied uniformly across tasks is: pixel observation wrapper; render resolution wrapper; crop wrapper; \texttt{FrameStack(3)}; \texttt{VecTransposeImage}; and SB3 image normalisation.

These experiments test whether replacing the standard image encoder with MiniConv preserves the ability to learn high-return behaviour under pixel observations. Within each task, MiniConv remains competitive with the Full-CNN baseline, but summary statistics exhibit task- and representation-size-dependent trade-offs between final and mean return. For each condition we report Best (maximum episodic return observed), Mean (average episodic return over training), and Final (mean episodic return over the final 100 episodes). These findings address Q1--Q2. Given that each condition is evaluated in a single fixed-seed run, the reported differences should be interpreted as indicative rather than statistically characterised.

\subsubsection*{Walker2d (PPO)}
Table~\ref{tab:walker2d:short} reports episodic return statistics for \emph{Walker2d} under PPO. The MiniConv encoder with $K=4$ attains a slightly higher final return than the Full-CNN baseline (3360 vs 3296), whilst the mean return over training is higher for Full-CNN (2800 vs 2680). The $K=16$ variant reaches the highest best episode but exhibits lower mean and final returns, suggesting less consistent performance under pixel observations.
\begin{table}[t]\centering
\caption{Walker2d (PPO): episodic return statistics over 2{,}000 episodes (single fixed-seed run).}
\label{tab:walker2d:short}
\begin{tabular}{lrrrr}\hline
\textbf{Architecture} & \textbf{Best} & \textbf{Final} & \textbf{Mean} & \textbf{Episodes} \\\hline
MiniConv encoder (K=4)  & 3640 & 3360 & 2680 & 2000 \\
MiniConv encoder (K=16) & 3800 & 3184 & 2320 & 2000 \\
Full-CNN                 & 3600 & 3296 & 2800 & 2000 \\
\hline\end{tabular}\end{table}

\subsubsection*{Hopper (SAC)}
Table~\ref{tab:hopper:short} reports results for \emph{Hopper} under SAC. The MiniConv encoder with $K=4$ yields the strongest final return, whilst Full-CNN attains the highest mean return. Across all encoders, the gap between best and mean/final indicates substantial variability in sustained performance under pixel observations. The relatively strong final return for $K=4$ is consistent with the possibility that a smaller representation encourages more robust behaviour, although we do not isolate this effect.
\begin{table}[t]\centering
\caption{Hopper (SAC): episodic return statistics over 2{,}000 episodes (single fixed-seed run).}
\label{tab:hopper:short}
\begin{tabular}{lrrrr}\hline
\textbf{Architecture} & \textbf{Best} & \textbf{Final} & \textbf{Mean} & \textbf{Episodes} \\\hline
MiniConv encoder (K=4)  & 2680 & 2360 & 1680 & 2000 \\
MiniConv encoder (K=16) & 2640 & 2200 & 1600 & 2000 \\
Full-CNN                 & 2656 & 2240 & 1720 & 2000 \\
\hline\end{tabular}\end{table}

\subsubsection*{Pendulum (DDPG)}
Table~\ref{tab:pendulum:short} reports results for \emph{Pendulum} under DDPG. Both MiniConv encoders outperform the Full-CNN baseline on final and mean return, with the $K=16$ encoder achieving the strongest overall performance (final return $-180$ vs $-248$ for Full-CNN).

This outcome is consistent with the character of the task: \emph{Pendulum} benefits from smooth, consistent control, and modest differences in representation quality can manifest as large differences in sustained performance. The improvement of $K=16$ over $K=4$ suggests that, for some tasks, a slightly richer transmitted representation can improve stability without substantially increasing the communication footprint.
\begin{table}[t]\centering
\caption{Pendulum (DDPG): episodic return statistics over 1{,}000 episodes (single fixed-seed run).}
\label{tab:pendulum:short}
\begin{tabular}{lrrrr}\hline
\textbf{Architecture} & \textbf{Best} & \textbf{Final} & \textbf{Mean} & \textbf{Episodes} \\\hline
MiniConv encoder (K=4)  & -140 & -192 & -244 & 1000 \\
MiniConv encoder (K=16) & -136 & -180 & -232 & 1000 \\
Full-CNN                 & -142 & -248 & -288 & 1000 \\
\hline\end{tabular}\end{table}

Taken together, these results suggest that MiniConv encoders can remain competitive with a conventional Full-CNN baseline under visual observations, but do not uniformly dominate across summary statistics. Encoder-$4$ achieves slightly higher final return on \emph{Walker2d} and \emph{Hopper}, whilst Full-CNN attains the higher mean return in both tasks; encoder-$16$ is less effective on the locomotion tasks but performs best on \emph{Pendulum}. This pattern indicates that the appropriate representation size is task-dependent and should be selected alongside device compute and bandwidth constraints.

\subsection{Execution Performance}

To characterise on-device feasibility, we measure per-frame inference time as a function of input size and device class, evaluate drift under sustained load, and record CPU temperature, RAM utilisation, and power consumption. These experiments address Q3--Q5, Q7, and Q8. We then analyse the computation--communication trade-off underpinning split inference to address Q6.

In addition to task-scale inputs, we include a high-resolution stress test (up to 3000$\times$3000) to expose throttling and power-limit behaviour under sustained load, particularly on the Jetson Nano.

\begin{figure}[t]
    \centering
    \begin{subfigure}{0.49\textwidth}
        \centering
        \includegraphics[width=\linewidth]{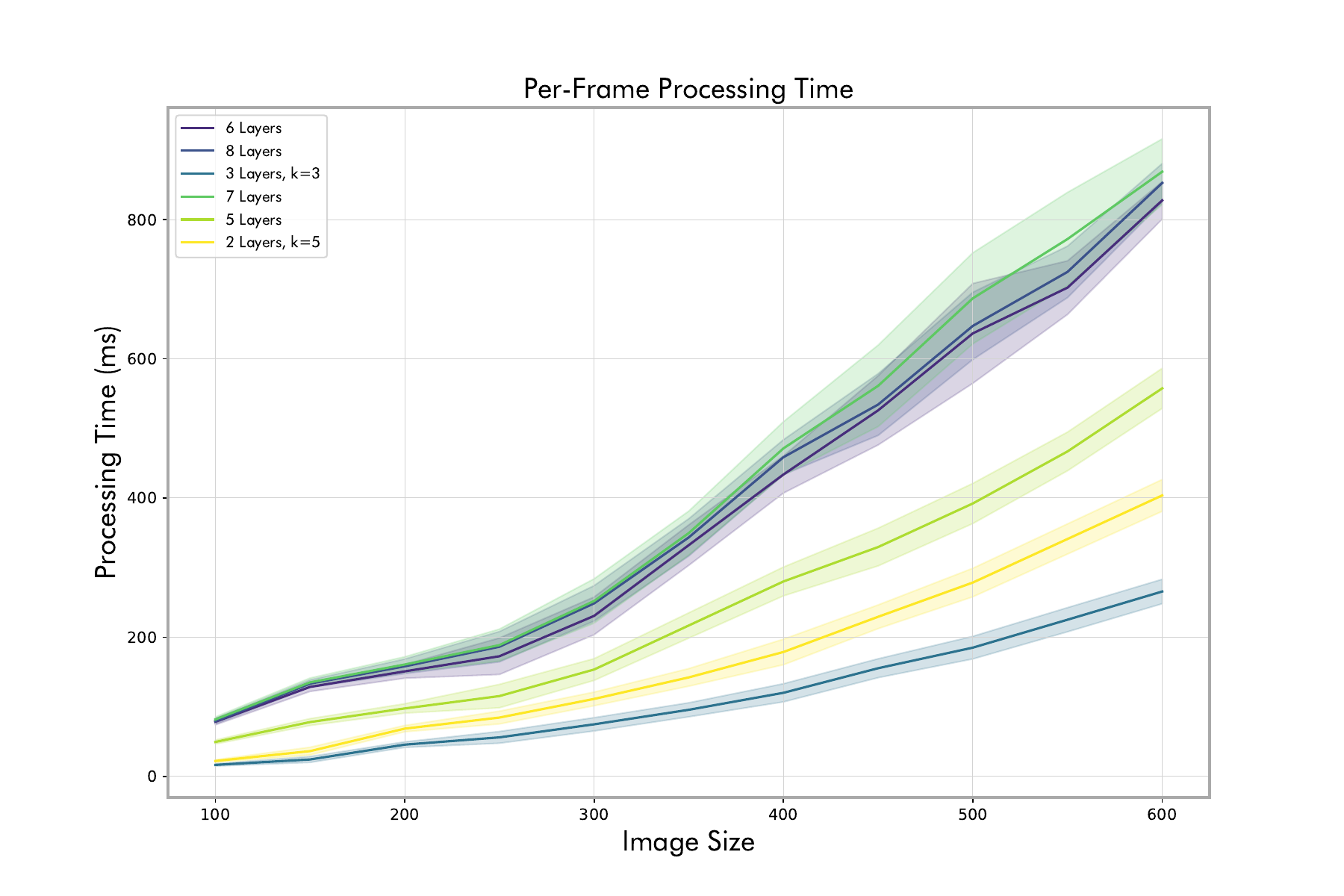}
        \caption{Raspberry Pi Zero 2 W.}
        \label{fig:execzero:short}
    \end{subfigure}
    \hfill
    \begin{subfigure}{0.49\textwidth}
        \centering
        \includegraphics[width=\linewidth]{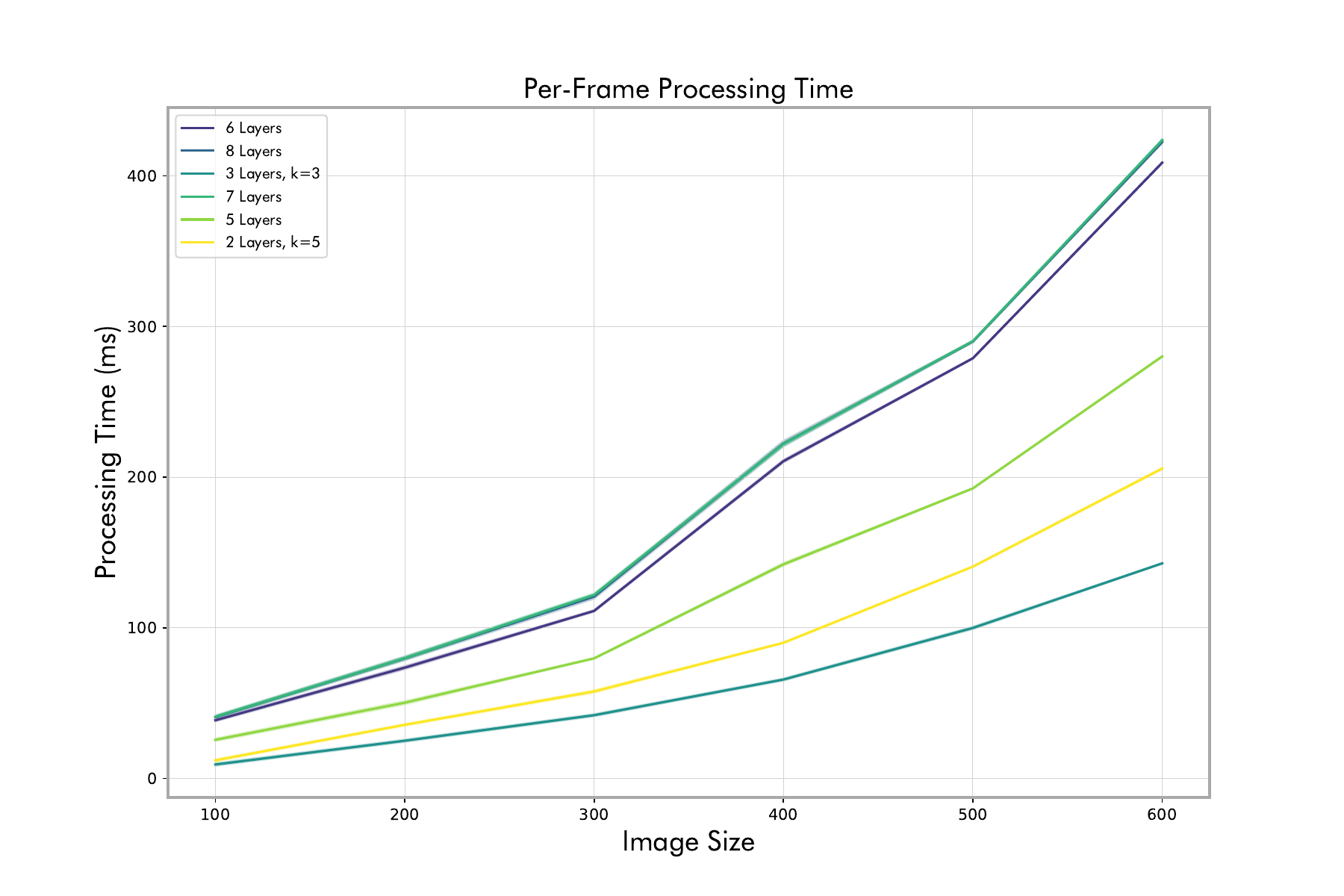}
        \caption{Raspberry Pi 4B.}
        \label{fig:exec4b:short}
    \end{subfigure}

    \medskip
    \begin{subfigure}{0.7\textwidth}
        \centering
        \includegraphics[width=\linewidth]{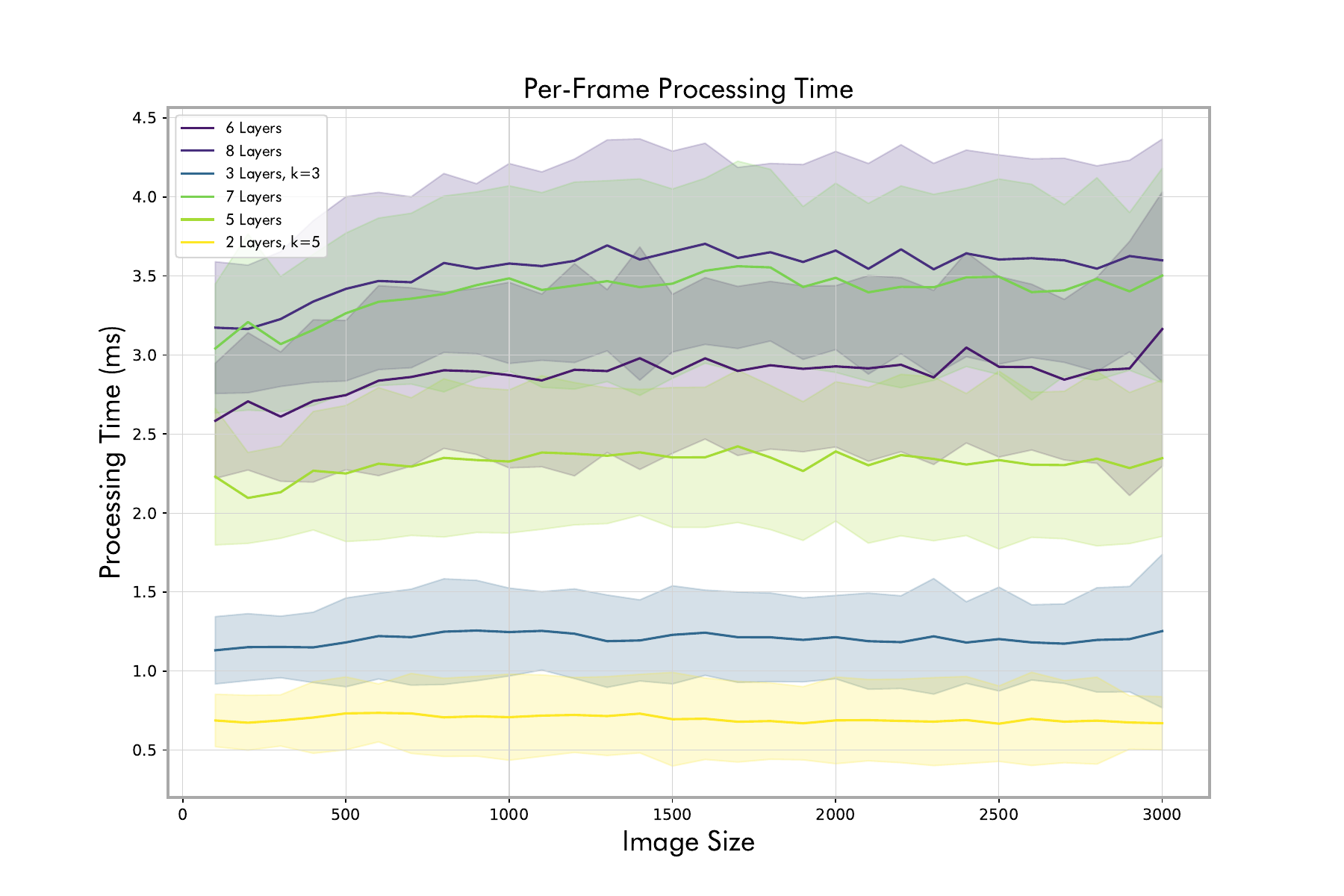}
        \caption{NVIDIA Jetson Nano.}
        \label{fig:execjetson:short}
    \end{subfigure}
    \caption{Per-frame processing time across devices as the input image size varies (mean of 100 consecutive inferences; shaded region shows standard deviation).}
    \label{fig:exec:short}
\end{figure}

Figure~\ref{fig:exec:short} summarises per-frame processing time across devices as the input size varies. As the input size increases, frame processing time increases on the Raspberry Pi platforms, whilst the Jetson Nano exhibits substantially lower times across the tested range. On the Pi Zero 2 W, maintaining a frame rate of five frames per second requires keeping the input size below $X=500$ (i.e., $500\times 500$).

\begin{figure}[t]
    \centering
    \begin{subfigure}{0.49\textwidth}
        \centering
        \includegraphics[width=\linewidth]{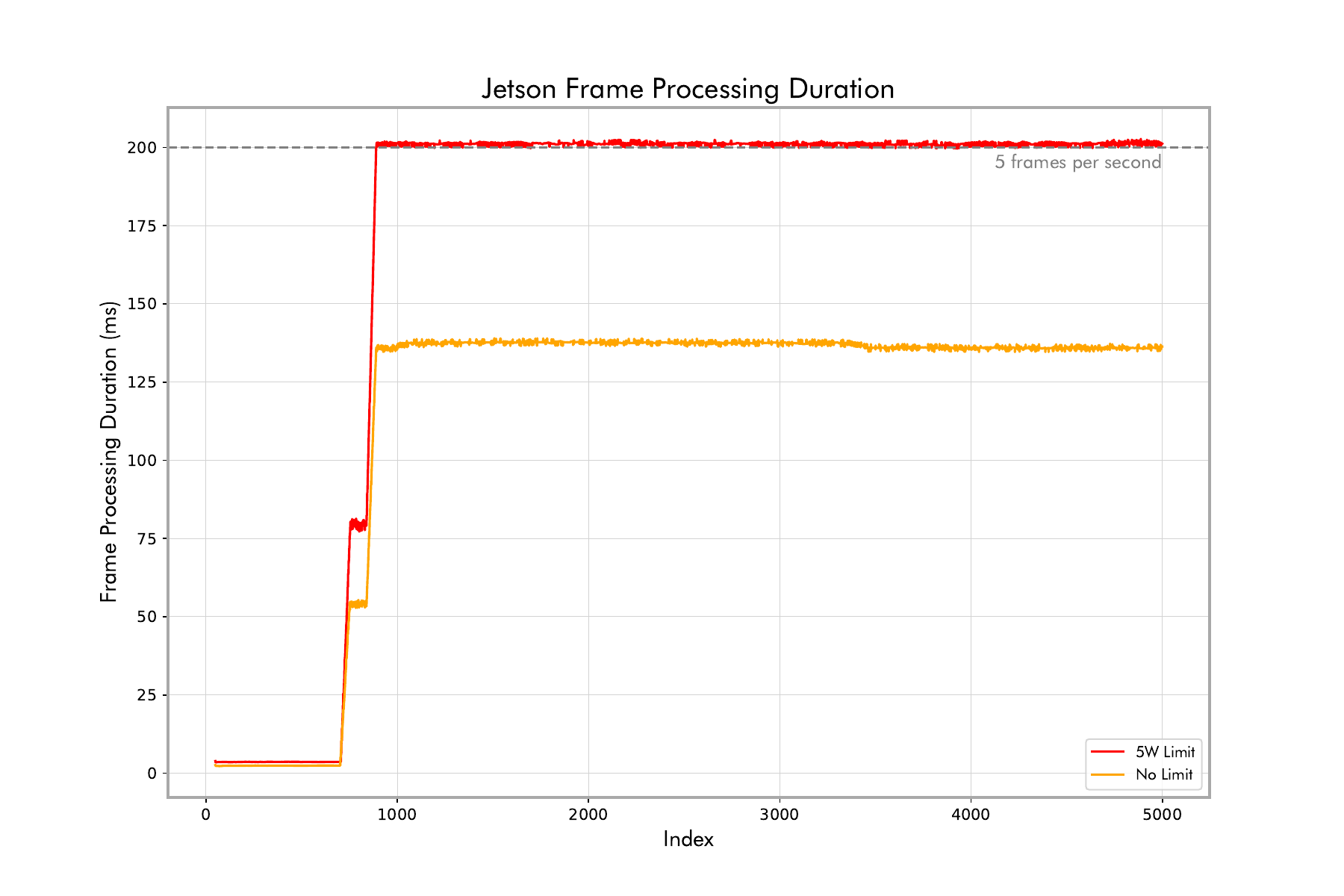}
        \caption{Jetson Nano (5W limit vs no limit).}
        \label{fig:jetsframe:short}
    \end{subfigure}
    \hfill
    \begin{subfigure}{0.49\textwidth}
        \centering
        \includegraphics[width=\linewidth]{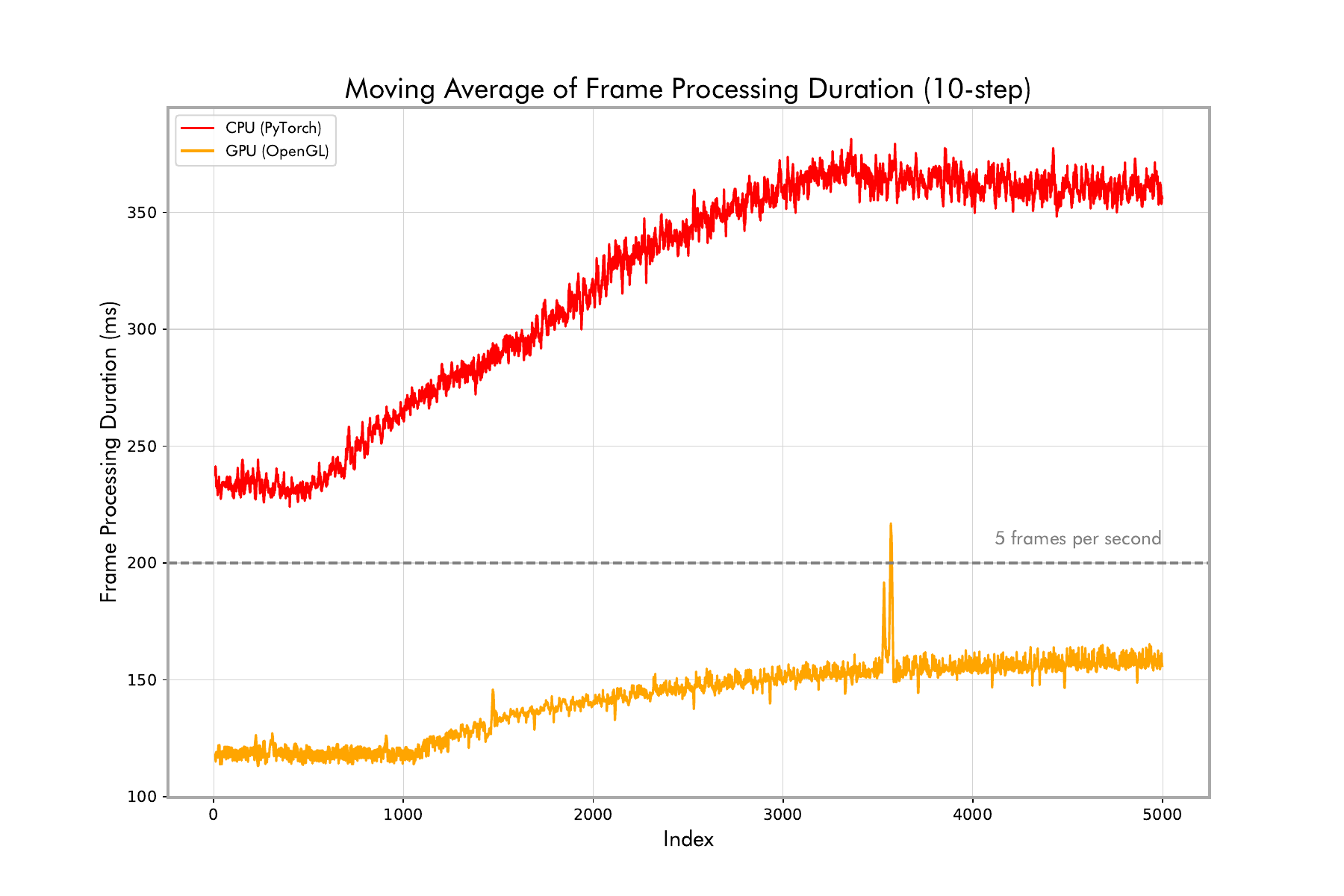}
        \caption{Pi Zero 2 W (moving average).}
        \label{fig:zerframe:short}
    \end{subfigure}
    \caption{Sustained inference performance over 5{,}000 consecutive frames.}
    \label{fig:sustained:short}
\end{figure}

To evaluate sustained performance, we measure inference time over extended runs. Figure~\ref{fig:sustained:short} shows that the Jetson Nano exhibits a marked increase in per-frame time after an initial period, and that power limits alter this behaviour. For the Pi Zero 2 W, GPU (OpenGL) inference is substantially faster and more stable than CPU (PyTorch) inference over the same horizon.

\begin{figure}[t]
    \centering
    \begin{subfigure}{0.49\textwidth}
        \centering
        \includegraphics[width=\linewidth]{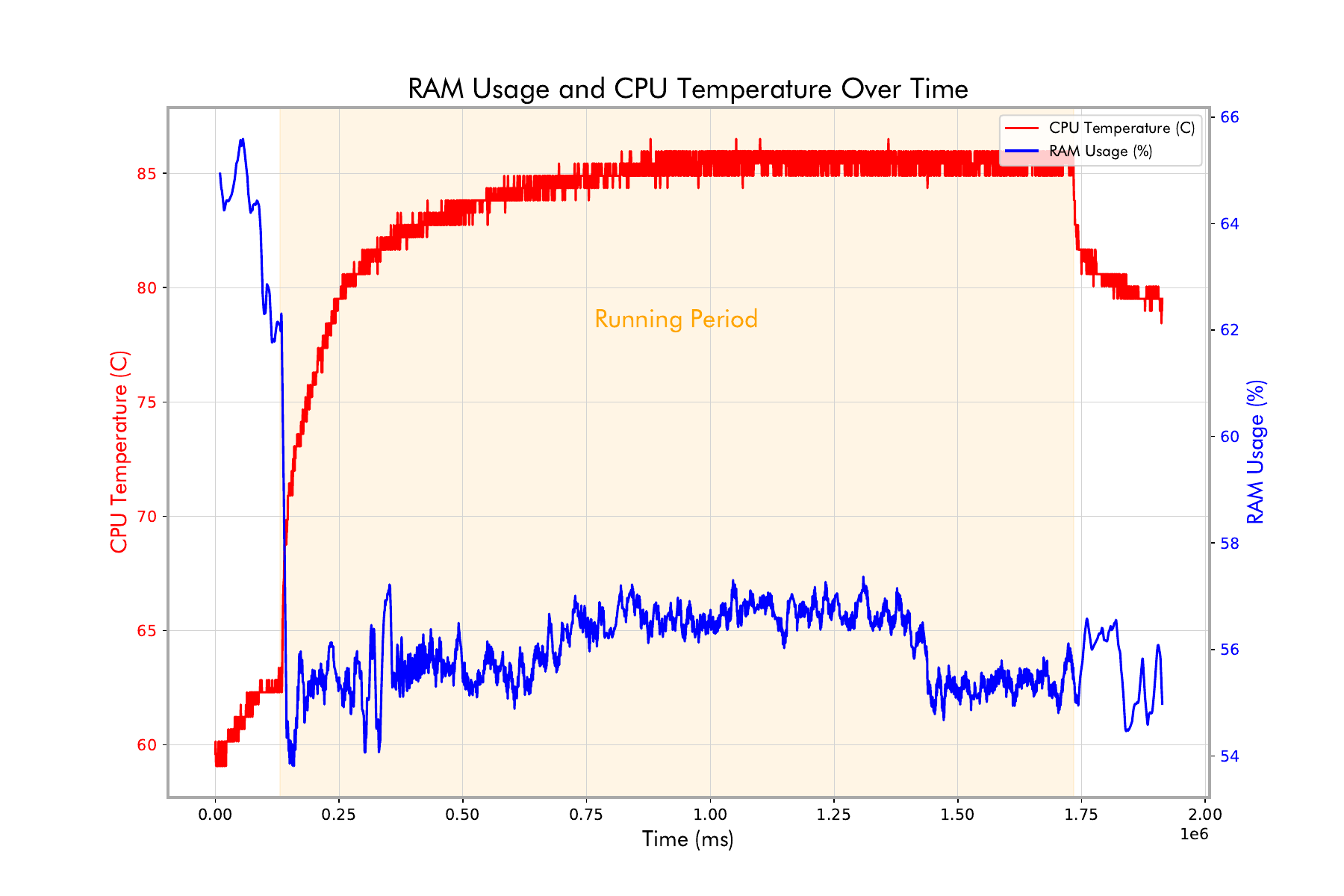}
        \caption{Pi Zero 2 W: CPU.}
    \end{subfigure}
    \hfill
    \begin{subfigure}{0.49\textwidth}
        \centering
        \includegraphics[width=\linewidth]{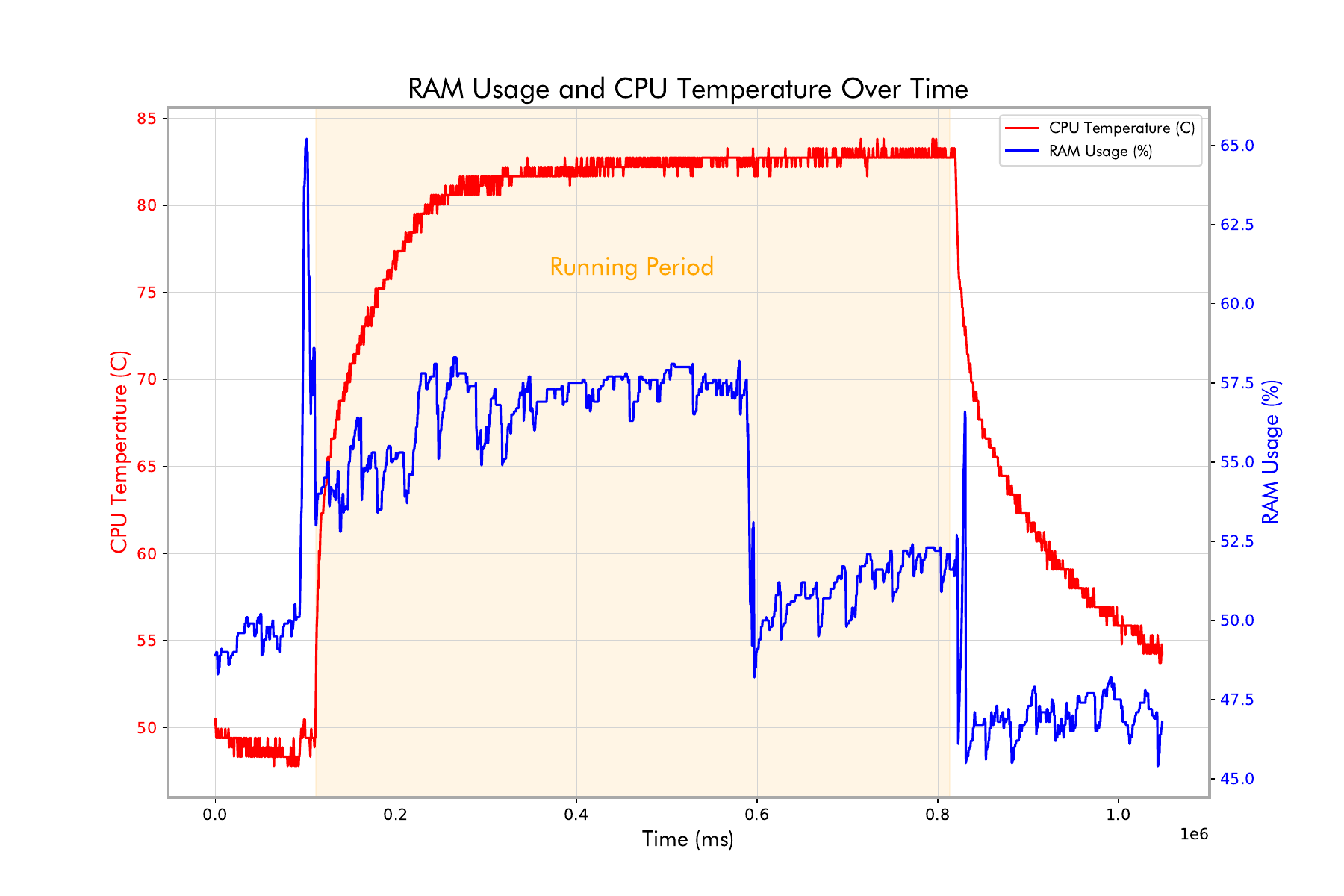}
        \caption{Pi Zero 2 W: GPU.}
    \end{subfigure}

    \medskip
    \begin{subfigure}{0.49\textwidth}
        \centering
        \includegraphics[width=\linewidth]{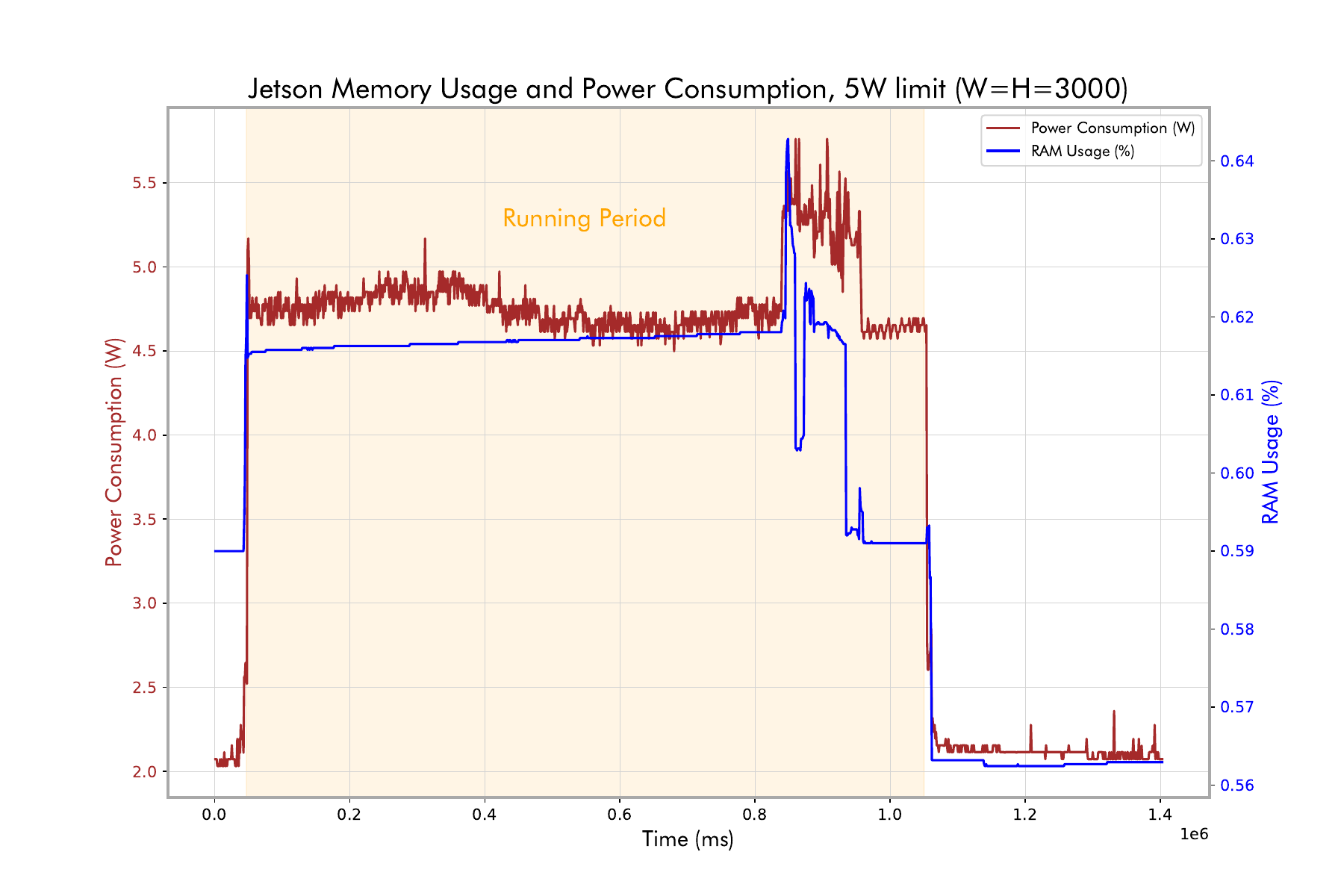}
        \caption{Jetson Nano: 5W limit.}
    \end{subfigure}
    \hfill
    \begin{subfigure}{0.49\textwidth}
        \centering
        \includegraphics[width=\linewidth]{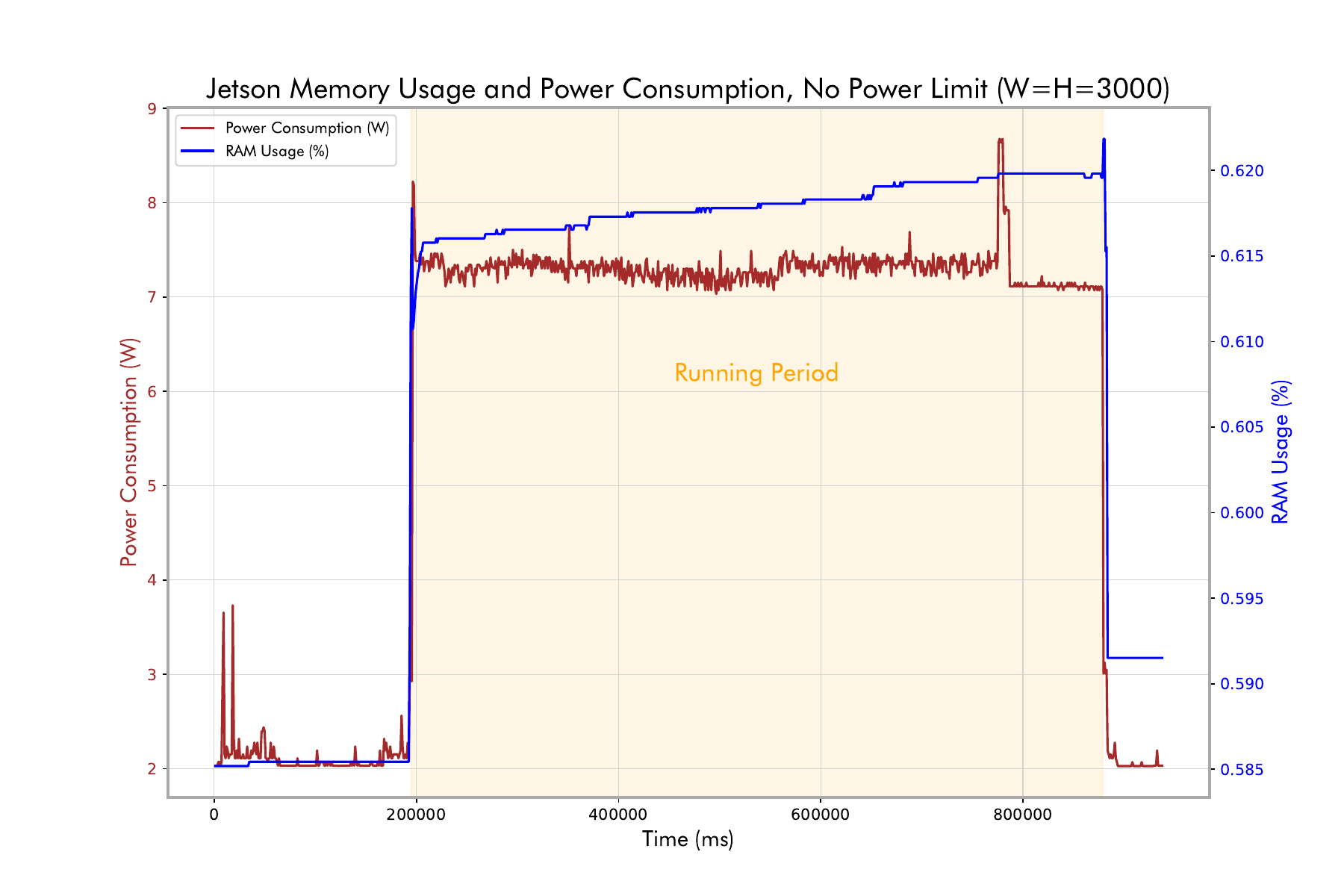}
        \caption{Jetson Nano: no power limit.}
    \end{subfigure}
    \caption{Resource usage during sustained inference (Pi Zero 2 W: RAM utilisation out of 512MB; Jetson Nano: power usage and memory pressure during 5{,}000 consecutive 3000$\times$3000 frames).}
    \label{fig:resources:short}
\end{figure}

To characterise the resource pressures associated with sustained inference, Figure~\ref{fig:resources:short} reports CPU temperature and RAM utilisation on the Pi Zero 2 W (CPU vs GPU execution), and power usage and memory pressure on the Jetson Nano (5W cap vs no limit). Across these experiments, RAM utilisation remains comparatively stable, whilst temperature and power reflect the expected constraints of sustained on-device execution.

\begin{figure}[t]
    \centering
    \includegraphics[width=0.9\linewidth]{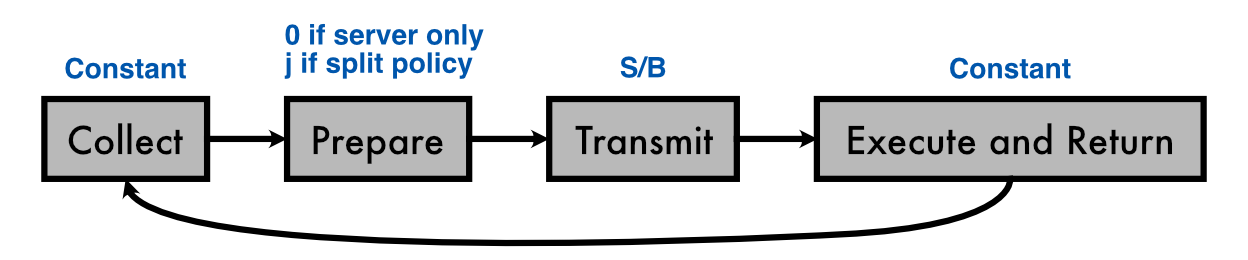}
    \caption{A breakdown of the steps involved in each decision that contribute to decision latency.}
    \label{fig:timesann:short}
\end{figure}

Ultimately, the utility of split-policy execution depends on the balance between computation and communication. Figure~\ref{fig:timesann:short} illustrates the decision-latency components that vary between a server-only pipeline and the split-policy pipeline.

We consider a simplified bandwidth model in which $B$ denotes link bandwidth in bits per second, $X$ denotes the input width and height, $n$ denotes the number of stride-two layers in the on-device encoder (so the transmitted feature map has spatial size $(X/2^n)\times(X/2^n)$), and $j$ denotes the per-frame on-device processing time. In our implementation, both raw observations and encoded features are transmitted as uncompressed uint8 buffers: a full RGBA frame requires $4X^2$ bytes, whilst a $K$-channel feature map requires $K(X/2^n)^2$ bytes (for the latency experiments we use $K=4$). Image compression would shift the break-even point and is left to future work. We ignore server-side compute to isolate the communication break-even point; server-side compute reductions are evaluated separately in the scalability experiment. Under these assumptions, split-policy inference yields a lower decision latency than a server-only pipeline when:
\begin{equation*}
    B < \frac{32X^2\left(1 - \frac{K}{4\cdot 2^{2n}}\right)}{j}.
\end{equation*}
For the Pi Zero 2 W configuration in Figure~\ref{fig:zerframe:short} ($X=400$, $n=3$, $j \approx 0.1s$, $K=4$), this yields a break-even bandwidth of approximately $50.4\,\mathrm{Mb\,s^{-1}}$.

\subsection{End-to-End Decision Latency}

To address Q6 empirically, we consider end-to-end \emph{decision latency}, measured as the median wall-clock time (over 1{,}000 decisions per setting) from the availability of an observation on the client device to the receipt of an action from the server. We compare a conventional client--server pipeline that transmits the full RGBA observation to the server against the split-policy pipeline, where the on-device encoder produces a spatially smaller $K=4$ representation and only this representation is transmitted.

Table~\ref{tab:e2e:short} summarises results under bandwidth shaping. At low bandwidth, the split-policy pipeline substantially reduces decision latency, as transmission dominates the decision loop. As bandwidth increases, the benefit diminishes and a crossover occurs, after which the additional on-device compute cost dominates.

\begin{table}[t]\centering
\caption{End-to-end decision latency under bandwidth shaping.}
\label{tab:e2e:short}
\begin{tabular}{lrr}\hline
\textbf{Bandwidth} & \textbf{Server-only latency (ms)} & \textbf{Split-policy latency (ms)} \\\hline
$10\,\mathrm{Mb\,s^{-1}}$  & 540 & 145 \\
$25\,\mathrm{Mb\,s^{-1}}$  & 240 & 140 \\
$50\,\mathrm{Mb\,s^{-1}}$  & 140 & 138 \\
$100\,\mathrm{Mb\,s^{-1}}$ & 90  & 137 \\
\hline\end{tabular}\end{table}

Consistent with the break-even analysis, the split-policy pipeline provides the largest reduction in decision latency at $10$--$25\,\mathrm{Mb\,s^{-1}}$, is approximately neutral around $50\,\mathrm{Mb\,s^{-1}}$, and becomes compute-bound on the client at higher bandwidth.

\subsection{Server Scalability}

A second practical motivation for the split-policy approach is to reduce the server-side compute cost per decision by moving the early visual feature extraction to the edge device. We consider a simple multi-client setting in which a single server processes requests from multiple concurrent clients, each operating at a fixed decision rate. Experiments are performed on a suitably powerful server with an Intel CPU and an NVIDIA GPU. Table~\ref{tab:scale:short} reports the maximum number of concurrent clients that can be supported at 10Hz whilst maintaining a p95 decision latency budget of 100ms.

\begin{table}[t]\centering
\caption{Server scalability at a fixed decision rate.}
\label{tab:scale:short}
\begin{tabular}{lrr}\hline
\textbf{Constraint} & \textbf{Server-only} & \textbf{Split-policy} \\\hline
10Hz per client, p95 latency $<100$ms & 12 clients & 36 clients \\
\hline\end{tabular}\end{table}

Under this simple setting, split-policy inference increases the number of concurrently served clients by approximately threefold under the same latency budget, reflecting the reduction in server-side compute per request. In practice, the achievable scaling depends on implementation details (e.g., batching and asynchronous I/O) and on the server hardware.

\section{Discussion}

By performing initial visual processing on-device, split-policy execution reduces the need to transmit raw frames, which can reduce exposure of sensitive information in camera and screen-based applications. However, compact feature representations can still leak information in principle; stronger privacy claims require explicit objectives or adversarial reconstruction testing. In deployment, standard transport encryption (e.g., TLS) remains necessary to protect transmitted features from third-party interception.

From a systems perspective, implementing the encoder as OpenGL shaders makes the approach compatible with a wide range of embedded GPUs and can reduce CPU memory traffic by processing frames within the graphics pipeline. The measurements also highlight a clear trade-off: at low bandwidth, communication dominates and split inference reduces decision latency; at higher bandwidth, the on-device encoder can dominate the decision loop. A limitation of the current learning evaluation is that results are reported for single fixed-seed runs; future work should characterise variance across seeds and environments, explore encoder designs and training objectives that improve the computation--communication trade-off, and extend evaluation to real camera pipelines and broader environments.

\section{Conclusion}

This paper introduced MiniConv, a library of small convolutional encoders designed to compile cleanly to OpenGL fragment shaders, and used it to realise a split-policy RL architecture in which early visual feature extraction is performed on-device. Across three visual control tasks, trained respectively with PPO, SAC, and DDPG, MiniConv encoders appear competitive with a conventional SB3 Full-CNN baseline under pixel observations. Encoder-$4$ achieves comparable final performance in these fixed-seed runs with modest differences in mean return, whilst encoder-$16$ exhibits mixed behaviour: weaker on the locomotion tasks but strongest on \emph{Pendulum}. These results suggest that representation size should be selected in a task-dependent manner.

The systems evaluation shows that the split-policy approach can substantially reduce end-to-end decision latency in bandwidth-limited settings (e.g., 540ms to 145ms at $10\,\mathrm{Mb\,s^{-1}}$) and can improve server scalability under a fixed latency budget in our testbed (12 to 36 concurrent clients at 10Hz, p95 $<100$ms). The central trade-off is explicit: benefits increase as bandwidth decreases and as the transmitted representation is made smaller, but additional on-device computation can dominate at higher bandwidth. Future work should refine encoder designs and training objectives to further improve this trade-off, and extend evaluation to real-world sensing pipelines and broader environments.

\section*{Code Availability}

The training scripts, on-device OpenGL deployment code, and measurement utilities described in this paper are available at \url{https://github.com/StandardRL-Components/MiniConv}. Measurement tooling for the device experiments is provided in two companion repositories: Jetson energy/telemetry collection \url{https://github.com/StandardRL-Components/JetsonMeasure} and a platform-agnostic performance harness \url{https://github.com/StandardRL-Components/SimplePerformanceMeasure}.

\clearpage
{\footnotesize
\begin{multicols}{2}
\bibliographystyle{IEEEtran}
\bibliography{refs}
\end{multicols}
}

\end{document}